\documentclass[letterpaper, 10 pt, conference]{ieeeconf}  % Comment this line out if you need a4paper

\IEEEoverridecommandlockouts                              % This command is only needed if 
\usepackage{amsmath} % assumes amsmath package installed
\usepackage{amssymb}  % assumes amsmath package installed
\usepackage{dsfont}
\usepackage{graphicx}

\usepackage{psfrag,graphicx,epsfig}
\usepackage{epstopdf}
\usepackage{xspace}
\usepackage{subfig}
\usepackage{float}
\usepackage{placeins}
\usepackage{multirow}
\usepackage{pgf,tikz}
\usepackage{nowidow}
\usepackage{lineno}
\usepackage{xcolor}
\newcommand{\fig}[1]{Fig.~\ref{#1}}
\newcommand{\tab}[1]{Table~\ref{#1}}
\newcommand{\eq}[1]{(\ref{#1})}
\DeclareMathOperator*{\argmin}{arg\,min}

\newcommand{\alg}[1]{Alg.~\ref{#1}}
\usepackage{siunitx}
\usepackage{color}
\usepackage{flushend}
\usepackage{lineno}
\usepackage{algorithm}
\usepackage[noend]{algorithmic}% Needed to meet printer requirements.
\allowdisplaybreaks

\title{\LARGE \bf
Covariance Steering for  Uncertain Contact-rich Systems}
\author{Yuki Shirai$^{\dagger}$, Devesh K. Jha$^{\ddagger}$, and Arvind U. Raghunathan$^{\ddagger}$%   
\thanks{$^{\dagger}$ Yuki Shirai is with the Department of Mechanical and Aerospace Engineering, University of California, Los Angeles, CA, USA 90095 {\tt\small yukishirai4869@g.ucla.edu}}%
\thanks{$^{\ddagger}$Devesh K. Jha and Arvind U. Raghunathan are with Mitsubishi Electric Research Laboratories (MERL), Cambridge, MA, USA 02139 {\tt\small \{jha,raghunathan\}@merl.com}}}%

\begin{document}

\maketitle

\begin{abstract}
 Planning and control for uncertain contact systems is challenging as it is not clear how to propagate uncertainty for planning. Contact-rich tasks can be modeled efficiently using complementarity constraints among other techniques. In this paper, we present a stochastic optimization technique with chance constraints for systems with stochastic complementarity constraints. We use a particle filter-based approach to propagate moments for stochastic complementarity system. To circumvent the issues of open-loop chance constrained planning, we propose a contact-aware controller for covariance steering of the complementarity system. Our optimization problem is formulated as Non-Linear Programming (NLP) using bilevel optimization. We present an important-particle algorithm for numerical efficiency for the underlying control problem. We verify that our contact-aware closed-loop controller is able to steer the covariance of the states under stochastic contact-rich tasks.
%  design trajectories with chance constraints for contact-rich tasks.
%  using the proposed closed-loop controller, we can achieve 
%  better performance when compared to the stochastic open-loop controller. 

%The proposed method is demonstrated for optimization in several contact-rich systems as well as a manipulation example with a physical 6 DoF manipulator arm. 

% Planning with uncertain complementarity systems is challenging as it is not clear how to propagate uncertainty 
% In this paper, we present a method for stochastic optimization in contact-rich systems with uncertainty. 

% Stochastic model predictive control (MPC) is a very powerful approach to perform control of uncertain systems. 

% However, stochastic MPC for contact-rich systems is challenging for nonlinear systems as exact uncertainty propagation is difficult for nonlinear systems. Furthermore, the solutions to stochastic MPC problems could be very conservative. In this paper, we present a technique for performing stochastic optimization in nonlinear systems using covariance steering. 
%     Our proposed method makes use of an exact method for uncertainty propagation and a feedback structure of the controller to allow better control of the uncertain nonlinear system. The proposed approach could also be useful for model-based reinforcement learning problems where the proposed controller design allows faster learning rates during learning. The proposed uncertainty aware optimization is demonstrated on several different systems as well as for a pendulum system during MBRL.
\end{abstract}
\section{Introduction}\label{sec:intro}
Contacts lead to discontinuous dynamics and thus, planning through contacts requires careful treatment of constraints arising due to these discontinuities. Complementarity constraints offer an efficient way of modeling contact systems. However, uncertainty in contact systems could lead to stochastic complementarity systems~\cite{shirai2022chance}. Even though complementarity systems are well studied, stochastic complementarity systems are not well understood. The state and complementarity variables are implicitly related via the complementarity constraints -- uncertainty in one leads to stochastic evolution of other. This makes uncertainty propagation challenging. %for the following stochastic optimization. 
Furthermore, multiplicity of solutions to the complementarity variables also makes it difficult to characterize the stochastic evolution. 
In this paper, we present an approximate treatment of stochastic complementarity systems using particles. We present the design and evaluation of a contact-aware stochastic controller for covariance control of the underlying uncertain system. An important-particle algorithm is presented for an efficient solution to the resulting stochastic optimization problem.

% We present a particle-based technique to perform feedback control of stochastic complementarity systems. We present covariance control for the stochastic complementarity systems by solving for a trajectory-centric feedback controller to enable efficient control along long-horizon trajectories. We present a cutting plane algorithm for computing numerically efficient solution to the proposed stochastic optimization problem. 
%Stochastic complementarity systems present unique challenges for control as it is difficult to propagate uncertainty.
%\djnote{Change the discussion from SMPC to chance constrained optimization}. \\
Chance-constrained optimization (CCO) has been extensively studied in the control of uncertain systems~\cite{4739221, blackmore2009convex, 5477242, 9143595, nakka2021trajectory, 9113247}. It allows us to plan using the uncertainty in the model by propagating the uncertainty which can be then used to design a controller for desired performance constraints of the system. However, in practice, the CCO techniques, \textcolor{black}{based on the analytical form of chance constraints}, impose restrictive assumptions of Gaussian uncertainty and linear constraints.
Further, state uncertainty increases with time and thus finding a controller for satisfying tighter state constraints could be infeasible over a long planning horizon. This is often the case in control of nonlinear systems with large uncertainty. This problem is aggravated for contact-rich systems due to the presence of discontinuities in system dynamics.
 
To circumvent these challenges, we consider particle-based method for uncertainty propagation and explicit covariance control of our contact-rich system during optimization.
%\textcolor{black}{Another challenge is the derivation of analytical expression for joint chance constraints over a trajectory (see \cite{prekopa2003probabilistic, doi:10.1137/070702928, 4739221, pagnoncelli2009sample, blackmore2009convex, shirai2022chance} for the discussion of joint chance constraints). The tractable method to deal with the joint chance constraints is based on Boole's inequality \cite{boole1847mathematical}, which introduces conservatism.}
% Motivated by the discussions above, we present a technique that tries to address them for stochastic contact-rich systems. First, we use particle-based uncertainty propagation to have the exact uncertainty propagation for the underlying dynamical system. Then we formulate the problem for covariance steering of the underlying system as a nonlinear optimization problem. 

\textbf{Contributions.} 
% The proposed work has the following contributions:
\begin{enumerate}
    % \item We present analysis of stochastic complementarity systems to show how uncertainty in state or complementarity variables affect each other.
    \item We present a novel formulation of covariance steering for complementarity systems using feedforward and feedback controller design.
    \item An important-particle algorithm is proposed for numerical efficiency and we evaluate the proposed method on several examples.
\end{enumerate}
While our motivation is to design robust feedback controllers for manipulation~\cite{9811812,https://doi.org/10.48550/arxiv.2303.08965}, the full problem is out of the scope of the current formulation. Thus, in this current paper, we limit the scope to linear complementarity systems with uncertainty.%, and evaluate the proposed approach on some academic examples. 
\section{Related Work}\label{sec:related_works}
% In this section, we present some work that is closely related to our proposed work. 
Our proposed stochastic optimization problem is mainly related to three major areas of work. The first major area is optimization with complementarity constraints. This topic has been well studied in optimization and robotics literature~\cite{posa2014direct, 9812069, 8740889, shirai_2022iros}. This approach has been shown to work well for generating trajectories for manipulation and locomotion problems. However, it cannot be trivially extended to stochastic complementarity systems to introduce robustness. More recently, contact-aware feedback controllers for contact-rich systems have been proposed \cite{9197568} for linear complementarity systems. However, it cannot also be extended to consider stochastic complementarity constraints to provide stochastic guarantees. 

Using stochastic complementarity constraints for planning robust manipulation is not so well understood in literature. Some of the recent work could be found in~\cite{drnach2021robust, shirai2022chance}. However, the problem with these approaches is that the uncertainty needs to be very small otherwise the optimization might be infeasible. Consequently, these approaches could fail to provide robust plans for uncertain contact systems. Furthermore, uncertainty propagation for stochastic complementarity systems is not properly modeled in these approaches. One of the reasons is the implicit relationship between contact and state variables in complementarity constraints. As a consequence, most of the known approaches (e.g., extended Kalman
filter \cite{thrun2002probabilistic}, unscented Kalman filter \cite{julier2004unscented},  moment-based \cite{wang2020fast, jasour2021moment}) for uncertainty propagation can not be used for stochastic complementarity systems.

% More recently, there has some to deal with long horizon planning for uncertain systems. 
Since open-loop CCO would lead to quite conservative solutions to satisfy chance constraints, 
  covariance steering methods have gained attention to deal with long-horizon planning for uncertain systems \cite{hotz1987covariance, okamoto2019optimal, ridderhof2019nonlinear}. Covariance steering methods are able to design feedforward and feedback gains simultaneously to satisfy chance constraints. However, these cannot be directly applied to contact-rich systems since they assume (in general) linear dynamics with Gaussian additive noises. 

In this paper, we present an approach for planning in stochastic contact-rich systems which formulates a robust controller design by considering covariance steering during planning. To understand uncertainty evolution in stochastic complementarity systems, we use particle-based control formulation~\cite{5477242, doi:10.1137/070702928, pagnoncelli2009sample, sehr2017particle} to get approximate uncertainty propagation. To the best of our knowledge, this is the first time that we have shown covariance steering with chance constraints for complementarity systems. 
\section{Problem Formulation}\label{sec:problem_statement}
In this section, we describe  preliminaries  of the method proposed in the current work.

\subsection{Stochastic Discrete-time Linear Complementarity Systems}\label{SDLCS_sec}
% In real world, there are many uncertainty sources such as modeling error and uncertain parameter identifications. To consider this effect, 
In this work, we consider the Stochastic Discrete-time Linear Complementarity Systems (SDLCS):
\begin{subequations}
\begin{align}
x_{k+1}=&{A}_k(\xi) x_k+B_k u_k+{C}_k(\xi) \lambda_{k+1}+{g}_k(\xi) \nonumber \\
&+ w_k(\xi) \label{slcp1} \\
0 \leq \lambda_{k+1} \perp& {D}_k(\xi) x_k+E_k u_k+{F}_k(\xi) \lambda_{k+1}+{h}_k(\xi) \nonumber \\
& + l_k(\xi) \geq 0 \label{slcp2}
\end{align}
\label{SDLCS_equations}
\end{subequations}
where $k$ is the time-step index, $x_k\in \mathbb{R}^{n_{x}}$ is the state, $u_k\in \mathbb{R}^{n_{u}}$ is the control input, and $\lambda_k \in \mathbb{R}^{n_{c}}$ is the algebraic variable (e.g., contact forces).
We define $x = [x_1, \ldots, x_T], u = [u_0, \ldots, u_{T-1}], \lambda = [\lambda_1, \ldots, \lambda_{T}]$. 
The parameter $\xi \thicksim \Xi$ is the uncertain parameter with distribution $\Xi$. In addition, ${A}_k(\xi) \in \mathbb{R}^{n_{x} \times n_{x}}$, $B_k \in \mathbb{R}^{n_{x} \times n_{u}}$, ${C}_k(\xi) \in \mathbb{R}^{n_{x} \times n_{c}}$, ${g}_k(\xi) \in \mathbb{R}^{n_{x}}$, ${D}_k(\xi) \in \mathbb{R}^{n_{c} \times n_{x}}$, $E_k \in \mathbb{R}^{n_{c} \times n_{u}}$, ${F}_k(\xi) \in \mathbb{R}^{n_{c} \times n_{c}}$, and ${h}_k(\xi) \in \mathbb{R}^{n_{c}}$ are all dependent on the uncertain parameter $\xi$.  For simplicity, we abbreviate $\xi$ from these matrices for the discussion in the following sections. 
% The $i$-th element of vector $p_k$ ($p_k$ can be $x_k, u_k, \lambda_k$) is represented as $p_{k,i}$.
% The $i$-th diagonal element of matrix $P_k$ is represented as $P_{k,ii}$.
The notation $0 \leq a \perp b \geq 0$ denotes the complementarity constraints $a \geq 0, b \geq 0, a b=0$. The initial state of the system $x_0(\xi)$ is also assumed to be uncertain. $\left\|x\right\|_Q^2$ means a quadratic term with a weighting matrix $Q$.

In the following, we make the assumption that $F_k(\xi)$ is a P-matrix~\cite{lcpbook} for all $k$ and $\xi$. Under this assumption, there is an unique solution $\lambda_{k+1}$ to~\eqref{slcp2} for each $\xi$ and any $u_k, x_k$.  From this it is easy to infer that there exists an unique trajectory $x$ and $\lambda$ for any realization of uncertainty $\xi \thicksim \Xi$ and controls $u$ from every initial condition $x_0(\xi)$.  In other words, we can define functions $\mathbf{x} : \Xi \times \mathbb{R}^{n_u(T-1)} \rightarrow \mathbb{R}^{n_xT}$ and $\boldsymbol{\lambda} : \Xi \times \mathbb{R}^{n_uT}$ that provides the unique trajectory given a realization of uncertainty, and the controls trajectory. Note that we do not show explicit dependence on initial condition due to the dependence of $x_0$ on the uncertain parameter $\xi$.

\subsection{Stochastic Control for Contact-Rich Systems}
% \textcolor{red}{YS: Maybe we can move this subsection to previous section and in this section we can focus discussing particle-baesd controller.} 
% In this work, we use particles to approximate the distribution of states and algebraic variables to design controllers for complementarity systems. First, we would like to formulate the general optimization problem:
In this work, we aim at finding a robust controller that satisfies chance constraints over SDLCS. To realize this, the following optimization problem can be formulated:
\begin{subequations}\label{equation_control}
\begin{flalign}
\min _{u}  &\;  \sum_{k=1}^{T}
\left\| \mathbb{E}_{\xi \thicksim \Xi} \left[\mathbf{x}_{k}(\xi,u)\right] - x_d \right\|_Q^2
% \left(\mathbf{x}_{k}(\xi,u) - x_d\right)^{\top} Q \left(\mathbf{x}_{k}(\xi,u) - x_g \right)
 + \sum\limits_{k=0}^{T-1}
\left\| u_k \right\|_R^2
% u_{k}^{\top} R u_{k} 
\label{exp_cost}\\
\text{s.t.}  &\; u_k \in \mathcal{U} \label{control_bnds} \\
&\; \text{Pr}_{\xi \thicksim \Xi}\left(\mathbf{x}(\xi,u) \in \mathcal{X}\right) \geq \Delta \label{chance_const}
\end{flalign}
\end{subequations}
where $Q=Q^{\top}$ is positive semidefinite, $R=R^{\top}$ is positive definite, $\mathcal{U}$ is a convex polytope consisting of a finite number of linear inequality constraints. 
$x_d$ is the target state at $t = T$.
The set $\mathcal{X}$ represents a convex safe region where the entire state trajectory has to lie in.  We assume that $\mathcal{X} = \{ x \in \mathbb{R}^{n_xT} \,|\, g_i(x) \leq 0 \,\forall\, i = 1,\ldots,n_g\}$. $\text{Pr}$ denotes the probability of an event and $\Delta$ is the user-defined minimum safety probability, where the probability of satisfying constraints is at least greater than $\Delta$.

%\begin{equation}
%    \mathbf{x}(\xi,u)\in \mathcal{X} \Longleftrightarrow \bigwedge_{m=1, \ldots, N_m} f_m(x_{1:T}) \leq 0
%\end{equation}
%where $\bigwedge_{m=1, \ldots, N_m} f_m(x_{1:T}) \leq 0$ represents $\mathcal{X}$. $N_m$ is the total number of inequality constraints associated with $\mathcal{X}$.
%Since $x_{1:T}$ is the trajectory of the random variable, we formulate chance constraints for $x_{1:T}$ over  $\mathcal{X}$  where Pr denotes the probability of an event. 

We propose to obtain an approximate solution to~\eqref{equation_control} using the Sample Average Approximation (SAA) introduced in~\cite{doi:10.1137/070702928, pagnoncelli2009sample}. We explain more details in Sec~\ref{sec:cov_control}. 

\section{Covariance Steering for Contact-Rich Systems}\label{sec:cov_control}

This section presents our proposed framework of stochastic optimal control for contact-rich systems. Our framework approximates the distribution of the state and algebraic variables  using particles. Under the assumption that $\bar{F}$ is P-matrix, our method can capture stochastic evolution of SDLCS such that we can formally guarantee the violation of states and design a closed-loop controller for SDLCS (i.e., covariance steering for SDLCS). 

%evolution of uncertainty through complementarity constraints by assuming  we can 

We first present our open- and closed-loop controller formulation for SDLCS using particles and  then present a computationally beneficial approach based on the active-point method \cite{jorge2006numerical} to accelerate the resulting optimization.

\subsection{Particle-based Control for Contact-Rich Systems}
% To overcome those issues, we propose to use particles to approximately represent the distribution of the state and algebraic variables.
We propose to solve~\eqref{equation_control} approximately using SAA by sampling the uncertainty.  In particular, we obtain $N$ realizations of the uncertainty $\Xi^N=\{\xi^1,\ldots,\xi^N\}$ by sampling the distribution $\Xi$.  In other words, we approximate the distribution $\Xi$ using a finite-dimensional distribution $\Xi^N$ which follows an uniform distribution on the samples. Accordingly, the SAA for~\eqref{equation_control} is given as 
\begin{subequations}\label{saa_equation_control}
    \begin{align}
        \min _{u}  &\;  \sum_{k=1}^{T} \left\|\mathbb{E}_{\xi \thicksim \Xi^N}\left[\mathbf{x}_{k}(\xi,u)\right] - x_d\right\|_Q^2  + \sum\limits_{k=0}^{T-1} \left\|u_k\right\|_R^2
        \label{saa_exp_cost}\\
        \text{s.t.}  &\; u_k \in \mathcal{U} \label{saa_control_bnds} \\
        &\; \text{Pr}_{\xi \thicksim \Xi^N}\left(\mathbf{x}(\xi,u) \in \mathcal{X}\right) \geq \Delta. \label{saa_chance_const}        
    \end{align}
\end{subequations}
Note that the distribution $\Xi$ has been replaced with the finite-dimensional $\Xi^N$ in the above to simplify the computation of the expectation in the objective and chance constraint.  However, there still remains the implicit function $\mathbf{x}(\xi,u)$ which requires us to simulate the SDLCS for every realization of $\xi \in \Xi^N$.  We opt to remove this difficulty by replacing the implicit functions with the corresponding trajectories $x^i, \lambda^i$ for each $\xi^i \in \Xi^N$.  

Our proposed computational formulation using $N$ particles is given by:
\begin{subequations}
\begin{flalign}
\min _{x^i, u, \lambda^i}   &\;  \sum_{k=1}^{T}
\left\| \frac{1}{N} \sum\limits_{i=1}^N x^i_{k} - x_d\right\|_Q^2
% (\bar{x}_{k} - x_g)^{\top} Q (\bar{x}_{k} - x_g)
+\sum_{k=0}^{T-1}\left\|u_k\right\|_R^2 
\label{exp_cost11}\\
\text{s.t.} &\; x_{k+1}^i={A}^i_k x_k^i+B_k u_k+{C}^i_k \lambda_{k+1}^i+{g}^i_k + w_k^i \label{slcp1_p111} \\
&\; 0 \leq \lambda_{k+1}^i \perp {D}^i_k x_k^i+E_k u_k+{F}^i_k \lambda_{k+1}^i \nonumber \\ & +{h}^i_k  +l_k^i \geq 0 \\
&\; x_0^i = x_0(\xi^i) \label{x0_condn} \\
&\; u_{k} \in \mathcal{U} \label{bounds_variables1}\\
&\; \frac{1}{N}\sum_{i=1}^{N} \mathbb{I}\left( x^i \in \mathcal{X}  \right) \geq \Delta \label{chance_const1}
\end{flalign}
\label{equation_control_MILP}
\end{subequations}
where $\mathbb{I}(\cdot)$ is an indicator function returning $1$ when the conditions in the operand are satisfied and $0$ otherwise. 
Note that $x^i, \lambda^i$ represent the state and algebraic variable trajectory, respectively, propagated from a particular set of particles $x_0^i,   \theta_k^i$ where $\theta_k^i = [{A}^i_k, {C}^i_k, {g}^i_k, {D}^i_k, {F}^i_k, {h}^i_k, w_{k}^i,  v_{k}^i]$. 
% Also, $x_{1:T}^i = [{x_{1}^i}^\top, \ldots, {x_{T}^i}^\top]^\top, \lambda_{0:T-1}^i = [{\lambda_{0}^i}^\top, \ldots, {\lambda_{T-1}^i}^\top]^\top$. 
Using $N$ trajectories obtained from $N$ particles, we approximate mean of random variables as 
$\mathbb{E}_{\xi \thicksim \Xi}[\mathbf{x}_k(\xi,u)] \approx \frac{1}{N} \sum_{i=1}^{N}x_k^i, \mathbb{E}_{\xi \in \Xi}[\boldsymbol{{\lambda}}_{k}(\xi,u)] \approx \frac{1}{N} \sum_{i=1}^{N}\lambda_k^i$. In \eq{equation_control_MILP}, we approximate \eq{exp_cost} using the mean variable as shown in \eq{exp_cost11}. 
Chance constraints \eq{chance_const} can be also approximated as \eq{chance_const1} using
$N$ realization trajectories, which can be formulated as integer constraints (see \cite{5477242}).

% In this paper, we 

In this work, we consider the following controllers:
\begin{subequations}
\begin{align}
 &\textbf{feedforward}: u_k = v_k \label{feedforward}\\
&\textbf{feedback}: u_k = v_k + K_k(x_k - \bar{x}_{k}) + L_k (\lambda_k - \bar{\lambda}_{k}) \label{feedback}
\end{align}
\end{subequations}
where $K_k, L_k$ are feedback gains to control covariance.
For brevity, we use $\bar{x}_k = \frac{1}{N} \sum_{i=1}^{N}x_k^i, \bar{\lambda}_k = \frac{1}{N} \sum_{i=1}^{N}\lambda_k^i$.
We emphasize that controlling both states and contact variables is critical for contact-rich systems and thus we also introduce $L_k (\lambda_k - \bar{\lambda}_{k})$ to \eq{feedback} to stabilize the system. 
Here, we focus on discussing feedback controller \eq{feedback} for \eq{equation_control_MILP}. The optimization formulation for covariance steering of SDLCS using particles would be: 
\begin{subequations}
\begin{flalign}
&\min _{x^i, v, K, L, \lambda^i}   \sum_{k=1}^{T}
||\bar{x}_{k} - x_d||_Q^2
% (\bar{x}_{k} - x_g)^{\top} Q (\bar{x}_{k} - x_g)
+\sum_{k=0}^{T-1}\left\|u_k\right\|_R^2   \label{exp_cost11_cov}\\
\text{s. t. } 
&x_{k+1}^i=({A}^i_k + B_k K_k) x_k^i+B_k v_k \nonumber \\ &+({C}^i_k + B_k L_k) \lambda_{k+1}^i + \bar{g}^i_k \nonumber \\
&-B_kK_k \bar{x}_k - B_kL_k \bar{\lambda}_{k+1} + w_k^i \label{slcp1_p111_cov}\ \\
&0 \leq \lambda_{k+1}^i \perp ({D}^i_k + E_k K_k) x_k^i \nonumber \\ &+E_k v_k+({F}^i_k + E_k L_k) \lambda_{k+1}^i \nonumber
\\
&+{h}^i_k -E_kK_k \bar{x}_k - E_kL_k  \bar{\lambda}_{k+1} + l_k^i \geq 0 \label{compl_milp} \\
&\eq{x0_condn}, \eq{bounds_variables1}, \eq{chance_const1} \label{const_milp}
\end{flalign}
\label{equation_control_MILP_cov}
\end{subequations}
% where $A^{cl, i}_k = {A}^i_k + B_k K_k,
% B^{cl}_k = B_k, 
% C^{cl, i}_k = {C}^i_k + B_k L_k, g^{cl, i}_k = \bar{g}^i_k -B_kK_k \mathbb{E}[x_k] - B_kL_k \mathbb{E}[\lambda_{k+1}],
% D^{cl, i}_k = {D}^i_k + E_k K_k, 
% E^{cl}_k = E_k, 
% F^{cl, i}_k = {F}^i_k + E_k L_k, h^{cl, i}_k = {h}^i_k -E_kK_k \mathbb{E}[x_k] - E_kL_k \mathbb{E}[\lambda_{k+1}]$. 
% 
% 
To solve \eq{equation_control_MILP_cov}, we need to take care of, \eq{slcp1_p111_cov}, \eq{compl_milp} and \eq{chance_const1}. One method is mixed-integer programming. It is possible that binary variables can be used to deal with integer constraints \eq{chance_const1} using Big-M formulation. 
Also, bilinear terms in \eq{slcp1_p111_cov} and \eq{compl_milp} can be approximated using McCormick envelopes, leading to additional binary variables. As a result, a number of binary variables are introduced and 
% where you introduce binary variables and use some methods such as Big-M formulation to handle integer constraints. However, 
we observed that it is almost impossible to obtain a single feasible solution.
Instead, in this work, we use NLP which can solve \eq{slcp1_p111_cov} as nonlinear constraints and  \eq{compl_milp} as complementarity constraints. We describe how we solve \eq{chance_const1} using NLP through complementarity constraints in Sec~\ref{bilevel_sec}.

% \textit{Remark 1}: 
% We can employ Mixed-Integer Linear Programming (MILP), similar to \cite{5477242}, to implement \eq{equation_control_MILP_cov} by replacing \eq{exp_cost11_} with linear costs. In order to use MILP for contact-rich systems, we need to make the following modifications. Firstly, we need to convert complementarity constraints  since they are nonlinear constraints. Secondly, we have other nonlinear terms such as $K_k x_k^i$. For the first case, we can use integer constraints such as big-M formulation to deal with complementarity constraints. For the second case, we can use McCormick envelopes to approximate bilinear terms. As a result, the MILP would introduce a number of integer variables and it is almost impossible to obtain a single feasible solution since the computational complexity is quite high. Thus, in this work, we decided to work on NLP based on bilevel optimization where it can only find locally-optimal controllers but at least we hope we can relatively quickly find solutions compared to the MILP method explained here. 

\subsection{Bilevel Optimization for Particle-based Control}\label{bilevel_sec}
To solve \eq{equation_control_MILP_cov} using NLP, we need to solve integer constraints \eq{chance_const1} in NLP fashion. To achieve this, we propose the following bilevel optimization problem. 
\begin{subequations}
\begin{flalign}
\min _{x^i, v, K, L, \lambda^i, t^i, z^*}   \sum_{k=1}^{T}
\left\|\bar{x}_{k} - x_d\right\|_Q^2
% (\bar{x}_{k} - x_g)^{\top} Q (\bar{x}_{k} - x_g)
+\sum_{k=0}^{T-1}\left\|u_k\right\|_R^2  \label{exp_cost11_bilevel}\\
\text{s. t. }  \eq{slcp1_p111_cov}, \eq{compl_milp}, \eq{bounds_variables1}\label{slcp1_p111_bilevel}\\
% \bigwedge_{m=1, \ldots, N_m} f_m(x_{1:T}^i)
\forall j = 1,  \ldots, n_g, \; g_j(x)\leq t^i,  \label{param_safe}\\
\frac{1}{N}\sum_{i=1}^N z^{i, *} \geq \Delta \label{approx_bilevel_chance}\\
\forall i = 1, \ldots, N, \; z^{i, *} = \argmin_{z^i} t^i z^i |0 \leq z^i \leq 1 \label{lower_opt}
\end{flalign}
\label{equation_control_bilevel}
\end{subequations}
We introduce time-invariant parameter $t^i \in \mathbb{R}^{1}$ for each set of trajectory realization $i$.  If  $x^i \in \mathcal{X}$, $t^i \geq -\epsilon$ with $\epsilon \geq 0$. In contrast, if  $x \not \in \mathcal{X}$, $t^i \geq 0$. This condition is encoded in \eq{param_safe}. 
We have in total $N$ lower-level optimization problems \eq{lower_opt}, where each optimization is formulated as linear programming.
% We solve linear programming for each lower-level optimization \eq{lower_opt},  where 
$z^i \in \mathbb{R}^{1}$ is the decision variable used in $i$~-th lower-level optimization problem. 

The purpose of \eq{lower_opt} is to count the number of trajectory realizations that are inside $\mathcal{X}$. The optimal solution of \eq{lower_opt} can be as follows:
\begin{equation}
 z^i=   \begin{cases}1, & t^i < 0 \\ \left[0, 1\right], & t^i = 0 \\0, &  t^i > 0 \end{cases}
\end{equation}
If $t^i < 0$, \eq{param_safe} argues that  $x^i \in \mathcal{X}$ and thus we count this $i$-th trajectory propagated from $i$-th particles as one. If $t^i = 0$, \eq{param_safe} argues $x^i \in \mathcal{X}$ ($x^i$ lies on the boundary of $\mathcal{X}$) and thus  we count this $i$-th trajectory propagated from $i$-th particles as one. If $t^i > 0$, then $x^i$ is not within $\mathcal{X}$, and thus we count it as zero.  Then \eq{approx_bilevel_chance} considers the approximated chance constraints. 
% Here, it is worth noting that our method considers joint chance constraints as showon in \eq{param_safe}. Thus, 

Since the upper-level optimization decision variable $t^i$ can be influenced by other upper-level decision variables, we need to solve these two problems simultaneously, leading to a bilevel optimization problem.
Since the lower-level optimization problems are formulated as $N$ linear programming problems, we can efficiently solve the entire bilevel optimization problem using the Karush-Kuhn-Tucker (KKT) condition as follows:
\begin{subequations}
\begin{flalign}
\min _{x^i, v, K, L, \lambda^i, t^i, z^{i, *}, w^i_+, w^i_-}  \eq{exp_cost11_bilevel} \label{exp_cost11_bilevel_MPCC}\\
\text{s. t. }  \eq{slcp1_p111_bilevel}, \eq{param_safe}, \eq{approx_bilevel_chance}\label{slcp1_p111_bilevel_MPCC}\\
\forall i = 1, \ldots, N, \; 0 \leq z^{i, *} \leq 1,  w^i_+, w^i_- \geq 0\label{lower_opt_MPCC}\\
w^i_+ (z^{i, *}-1) = 0, w^i_- (z^{i, *}) = 0, \\
t^i + w^i_+  - w^i_-  = 0
\end{flalign}
\label{equation_control_bilevel_MPCC}
\end{subequations}
where $w^i_+, w^i_-$ are Lagrange multipliers associated with $z^i - 1 \leq 0$, $-z^i  \leq 0$, respectively. In conclusion, we obtain a single-level nonlinear programming problem with complementarity constraints, which can be efficiently solved using an off-the-shelf solver such as IPOPT \cite{80fe29bf9dc245ffa5c8bd7b3eee2902}.

% \textit{Remark:} $t^i$ can take zero when the trajectory $x_{1:T}^i$ lies on the edge of $\mathcal{X}$. In this case, $z^i$ can take any arbitrary value from $[0, 1]$. Hence, \eq{lower_opt} might not correctly count the number of trajectories which are on $F$. However, if the solver finds a feasible solution, it means that the solver can satisfy all constraints including \eq{approx_bilevel_chance}. Therefore, it means that the solver could find a feasible trajectory with chance constraints.
% Another way of thinking of this problem is as follows. $t^i=0$ means that $x_{1:T}^i$ lies on the edge of $F$, which is feasible. Thus, $z^i$ can take any value given constraints $0 \leq z^i \leq 1$ to satisfy \eq{approx_bilevel_chance}. [Need to be re-written]

\subsection{Important-particle Method for Particle-based Control}
One limitation of our method in Sec~\ref{bilevel_sec} is that the computation can be demanding with many particles to capture the evolution of uncertainty.
% To capture the evolution of uncertainty well, a number of particles might be necessary, which 
In this section, we present an approximate algorithm which samples important particles which might be most informative for constraint violation.
%We are interested in if we really need to have many particles to train the controller. 
To decrease the computational burden, we employ an important-particle method (see \alg{cutting_plane}) which starts from a relatively small number of particles and keeps adding  particles if the chance constraints are not satisfied due to the lack of the accurate approximation of variables. Since we start from a small number of particles, it is possible that our optimization could quickly find a feasible solution which works over testing data set. However, in the case when the problem is infeasible for some particles, we add the particles which experience maximum constraint violation to our set. Thus, we call our proposed method "important-particle" method-- the worst particles specify the boundary of feasible sets.

%If the problem is really complicated and the uncertainty is quite high, it might not happen but we still hope that  the performance of our controller improves monotonically by adding the "bad" particles, which make the controller fail to capture the evolution of uncertainty. That's why we call our method as "active-point" method - the worst particles specify the boundary of feasible sets. 
% We hope our cutting plane method would work better than the naive optimization method \eq{equation_control_bilevel_MPCC} since each iteration 

The pseudocode of our important-particle method for covariance steering is shown in \alg{cutting_plane}. 
% Given $\alpha$ particles for training the controller and $\beta$ particles for testing the controller, our method 
Param is the collection of parameters such as $Q, R$. $\alpha, \beta$ represent the number of particles for training and testing the controller, respectively. $\gamma$ is the number of initial particles our method uses during its first iteration. $\eta$ is the number of particles our methods adds to \eq{equation_control_bilevel_MPCC} for each iteration. 

As shown in \alg{cutting_plane}, our method keeps adding more particles unless either it runs more than MAX-ITER or converges to user-defined $\Delta$ given threshold $\Delta_\text{th}$. For each iteration, we run \eq{equation_control_bilevel_MPCC}. If the obtained solution is feasible, we do Monte Carlo simulation (MC simulation) over the training data set with $\alpha$ particles and calculate the empirical safe probability $\Delta_\alpha$. If this $\Delta_\alpha$ is close to or greater than $\Delta$, we terminate the while loop and run the obtained controller over the testing data set with $\beta$ particles. Otherwise, we choose the $\eta$ worst particles based on how much they violate the chance constraints and add them to $\theta$. If we obtain the infeasible solution or the "restoration phase failed" solution in IPOPT, we randomly choose the $\eta$ particles.

  \begin{algorithm}[t]
    \small 
  \algsetup{linenosize=\small}
 \caption{$\operatorname{ImportantParticle}(\text{Param}, \alpha, \beta, \gamma, \eta)$}
 \label{cutting_plane}
 \begin{algorithmic}[1]
 \STATE $j = 0$, $\theta = \gamma$, $\Delta_\alpha = 0$
 \WHILE{$j\leq \text{MAX-ITER}$ \AND $(\Delta-\Delta_\alpha)^2 \geq \Delta_\text{th}$ \AND $\Delta > \Delta_\alpha$;} 
 \STATE Run \eq{equation_control_bilevel_MPCC} with $N=\theta$
  \IF{The obtained solution from \eq{equation_control_bilevel_MPCC} is feasible}\label{4400}
 \STATE Run MC simulation with $\alpha$ particles and calculate $\Delta_\alpha$.\label{line23}
 \STATE Choose the $\eta$ worst particles that violate chance constraints.
 \ELSE
 \STATE Choose the random $\eta$ particles.
 \ENDIF
 \STATE $\theta = \theta + \eta$
 \ENDWHILE
 \STATE Run MC simulation with $\beta$ particles and calculate $\Delta_\beta$.
  \RETURN $x^{i, *}, v^*, K^*, L^*, \lambda^{i, *}, t^{i, *}, z^{i, *}, w^i_+, w^i_-, \Delta_\beta$
 \end{algorithmic} 
 \end{algorithm}

\section{Results}\label{sec:results}
In this section, we present  numerical results for our proposed approach and compare  them against some baselines.
%We present  numerical results to highlight the problems with different assumptions and approaches for uncertain contact-rich systems and how these could be overcome by the approach presented in the paper. 
% Since stochastic complementarity systems are not so well studied in literature, we start with presenting some basic analysis along with results of the proposed controller. 
In particular, we would like to highlight and understand the following questions:
\begin{enumerate}
    \item Does uncertainty in complementarity constraints lead to uncertainty in state trajectory?
    % \item Can ERM sufficiently model evolution of state and complementarity variables for SDLCS?
    \item How does the proposed controller perform of variance of states for SDLCS?
    % \item How well the proposed controller perform when compared to a stochastic open-loop controller?
\end{enumerate}
% Analysis of the first two questions provides an understanding of the challenges for controlling uncertain contact-rich systems. In the remaining two questions, we propose a solution that can potentially alleviate some of these issues.

% To answer the above four questions, we plan to show the following results:
% \begin{enumerate}
% \item Time history of $\lambda, y$ and time history of $x$: Then, we can show that $x_{k+1}$ has distribution if $\lambda_{k+1}$ has distribution. Answering question No. 1. We can also show the opposite direction as well.  
% \item Show MC results with ERM. Then we can show that controller could not stabilize the system since uncertainty evolution from complementarity constraints to state dynamics is not captured even if optimization can find feasible solutions. 
% \item Show figures for MC results with mean planned trajectory with chance constraint bounds. Show table that summarizes chance constraints satisfaction. Also, to show chance constraints are active, show MC results with different probability. Open-loop controller will be also shown. 
% \end{enumerate}

% \subsection{Experiment Setup}
We implement our method using IPOPT \cite{80fe29bf9dc245ffa5c8bd7b3eee2902} with PYROBOCOP \cite{9812069}. The optimization problem is implemented on a computer with Intel i7-12700K processor. 
We set $\alpha=250, \beta=1000$ for \alg{cutting_plane}. For $\gamma$ and $\eta$ in \alg{cutting_plane}, we use the different values for different applications as shown in \tab{table_cartpole} and \tab{table_acrobot}. When we run \eq{equation_control_bilevel_MPCC} alone without using  \alg{cutting_plane}, we use 1000 samples to calculate the empirical probability of failure to evaluate the satisfaction of chance constraints. 

% To verify the robustness of open-loop trajectories obtained from our proposed optimization, we use MC simulations.
Here we explain how we simulate trajectories (i.e., perform MC simulation for SDLCS, see~\cite{shirai2022chance} for more details).
\textcolor{black}{We propagate the dynamics by finding the roots of the complementarity system with sampled parameters given the control sequence obtained from optimization. We run each case for 1000 trials with different sampled parameters to estimate the probability of failure.}
 Note that, unlike the continuous-domain dynamics, we cannot rollout the dynamics for SDLCS with the given control sequences since we do not have the access to $\lambda_{k+1}$. 
%  We add the noise sampled from the distribution which was used during optimization.

\subsection{Uncertainty Propagation for SDLCS}\label{uncertainty_demo_SDLCS}

\begin{figure}[t]
    \centering
    \includegraphics[width=0.25\textwidth]{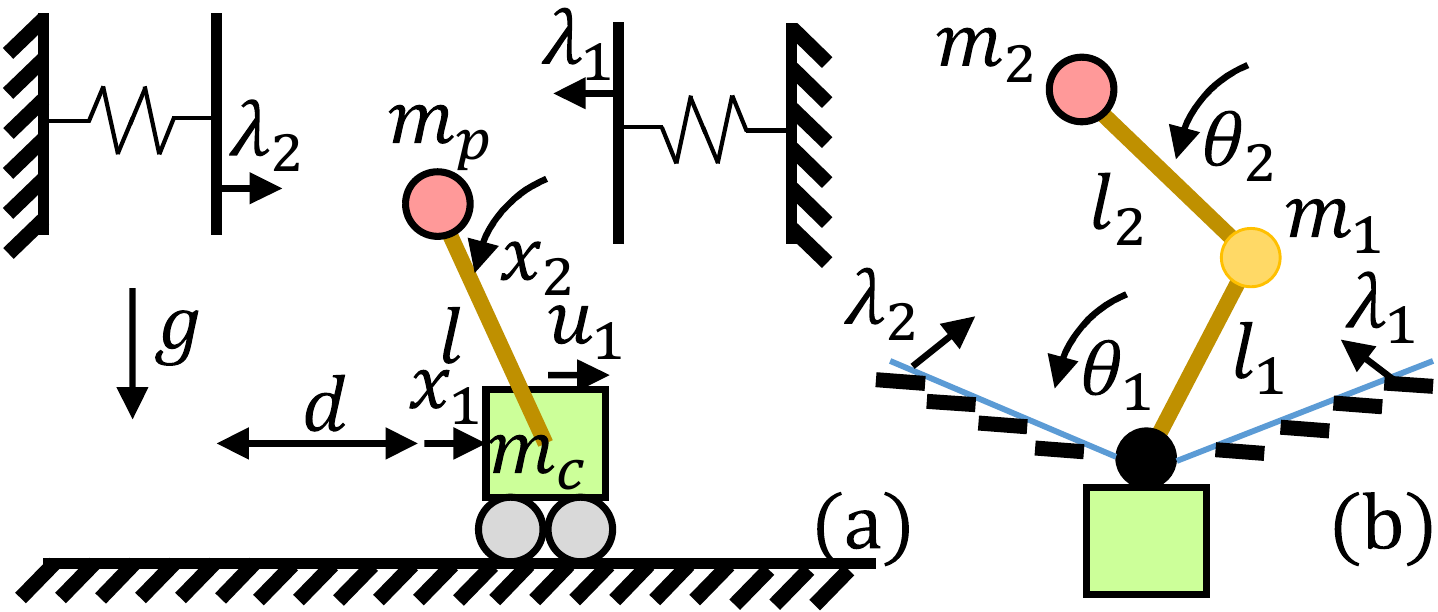} %
    \caption{(a): cartpole with softwalls. (b): acrobot with soft joints.}
    \label{fig:dynamics}
\end{figure}

\begin{figure}[t]
    \centering
    \includegraphics[width=0.499\textwidth]{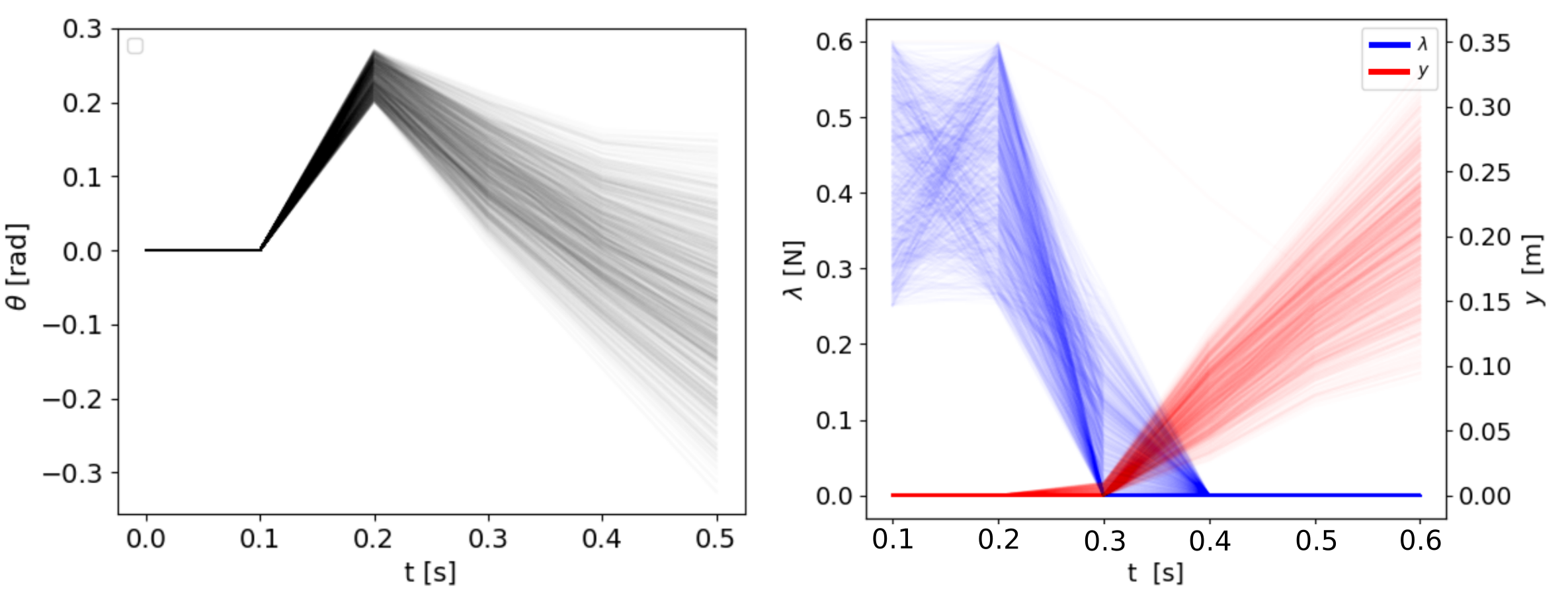} %
    \caption{Uncertainty propagation for cartpole system. Here only uncertainty arises from stiffness parameters $k_1, k_2$. }
    \label{fig:q1}
\end{figure}

\begin{figure}[t]
    \centering
    \includegraphics[width=0.479\textwidth]{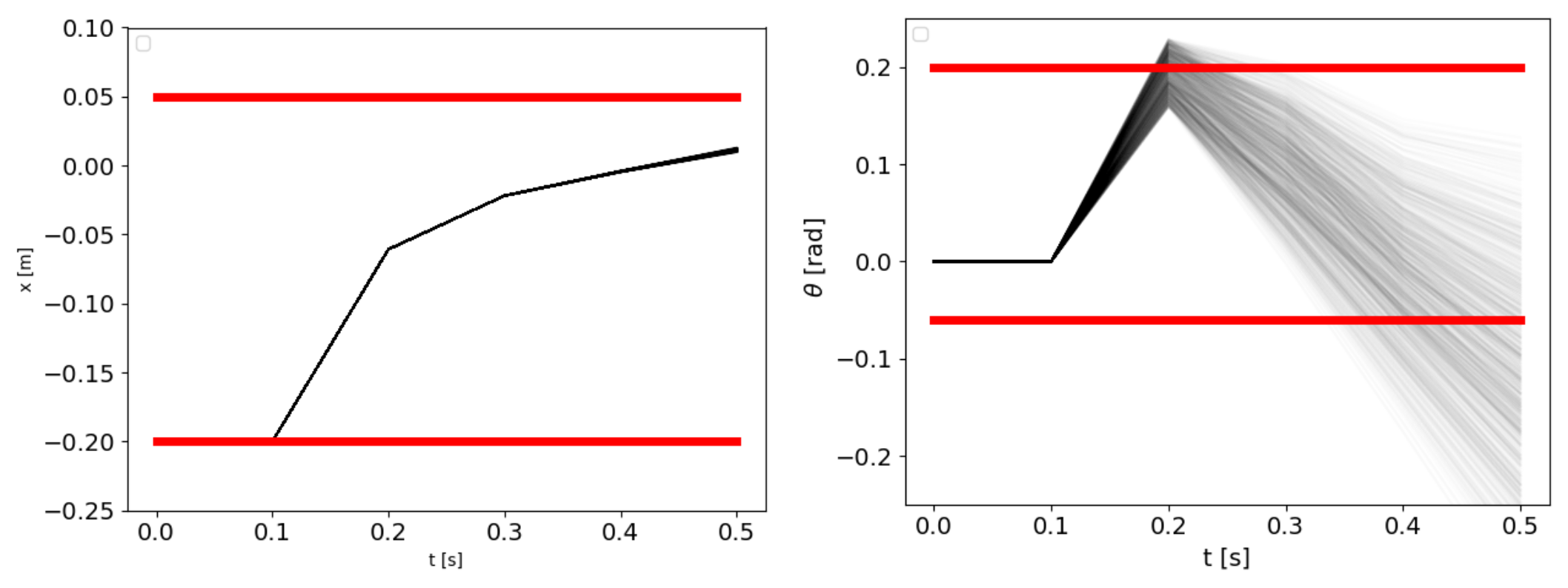} %
    \caption{Simulated trajectories for cartpole system using ERM-based controller. $\Delta=0.2$ and $\Delta_\text{test} = 0.083$. Red lines show boundaries specified in chance constraints.}
    \label{fig:cartpole_ERM}
\end{figure}

We show uncertainty evolution for SDLCS. We demonstrate this for a cartpole system with softwalls (see \cite{9197568} for more details). Here we consider both $k_1$ and $k_2$ follows uniform distributions where upper bound of uniform distribution for $k_1$ and $k_2$ is 14, 12, respectively, and the lower bound is 5 for both $k_1$ and $k_2$. In this experiment, we do not run any controller: we simply propagate SDLCS given uncertain parameters in order to show how the SDLCS behaves.

\fig{fig:q1} shows the evolution of uncertainty for the aforementioned system. At $t=0$ s,  there is no uncertainty for state $\theta_{t=0}$. However, because we provide uncertainty with $k_1$ and $k_2$, $\lambda_{t=0.1}$ has uncertainty. This is again because given realization of uncertain parameters, complementarity constraints give a realization of $\lambda$ and $y$, resulting in uncertainty in $\lambda$ and $y$. This stochastic $\lambda_{t=0.1}$ brings uncertainty in $\theta_{t=0.1}$ based on \eq{SDLCS_equations}. As shown in \fig{fig:q1}, both state and complementarity variables are stochastic. This can not be captured in approximations like Expected Residual Minimization (ERM)~\cite{drnach2021robust}.
% \djnote{Point out the distribution of different variables as a function of time. This should lead to the discussion as in why the previous approaches can fail.}

\subsection{Cartpole with Softwalls}
% We demonstrate our work for cartpole with softwalls system (see \cite{9197568} for more details). 
% \djnote{Point out the difference between contact-aware and the non-contact-aware controllers.}
% Finally, we discuss the difference between our proposed contact-aware and the non-contact-aware (i.e., $L_k=0, \forall k$ in \eq{feedback}). To observe how controllers behave in a longer horizon, we set $T = 20$ and use the same values for the other parameters. We observed that  
% Therefore, we conclude that introducing feedback to both states and forces is important to design feedback controller over SDLCS. 
We demonstrate our open- and closed-loop controllers for cartpole with softwalls system. 
$x$ is the cart position and $\theta$ is the pole angle. $u_{1}$ is the control and $\lambda_{1}, \lambda_{2}$ are the reaction forces at from the wall 1, 2, respectively.
We have the following deterministic physical parameters. $g=9.81$ is the gravitational acceleration, $m_p=0.1, m_c=1.0$ are the mass of the  pole, cart, respectively. $l=0.5$ is the length of the pole and $d=0.15$ is the distance from the origin of the coordinate to the walls.  
We assume that the uncertainty arises from the ${k_1}, {k_2}$ and use the same distribution in Sec~\ref{uncertainty_demo_SDLCS}. We set $dt=0.1$ for the explicit Euler integration and $T=6$.
\begin{figure}[t]
    \centering
    \includegraphics[width=0.479\textwidth]{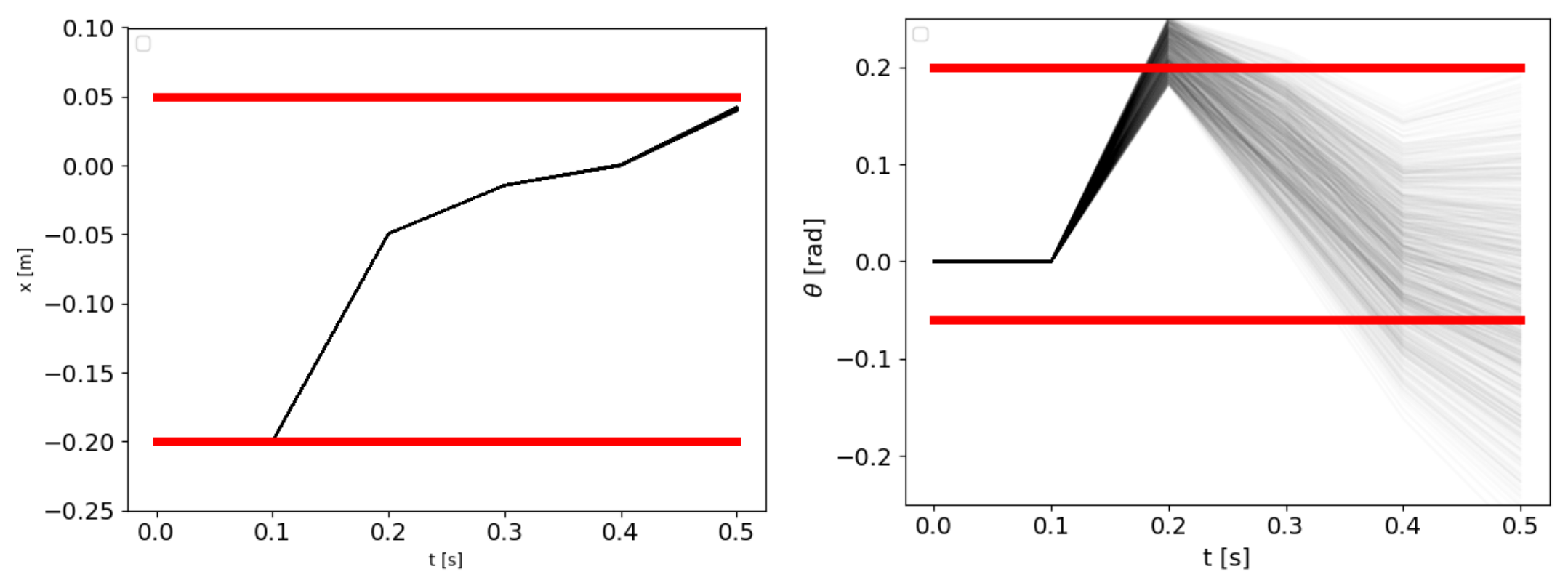} %
    \caption{Simulated trajectories for cartpole system using our open-loop controller. $\Delta=0.2$ and $\Delta_\text{test} = 0.190$ where $\Delta$ is input of optimization and $\Delta_\text{test}$ is the empirically obtained success rate from MC simulation. Red lines show boundaries specified in chance constraints.}
    \label{fig:cartpole_open}
\end{figure}

\begin{figure}[t]
    \centering
    \includegraphics[width=0.479\textwidth]{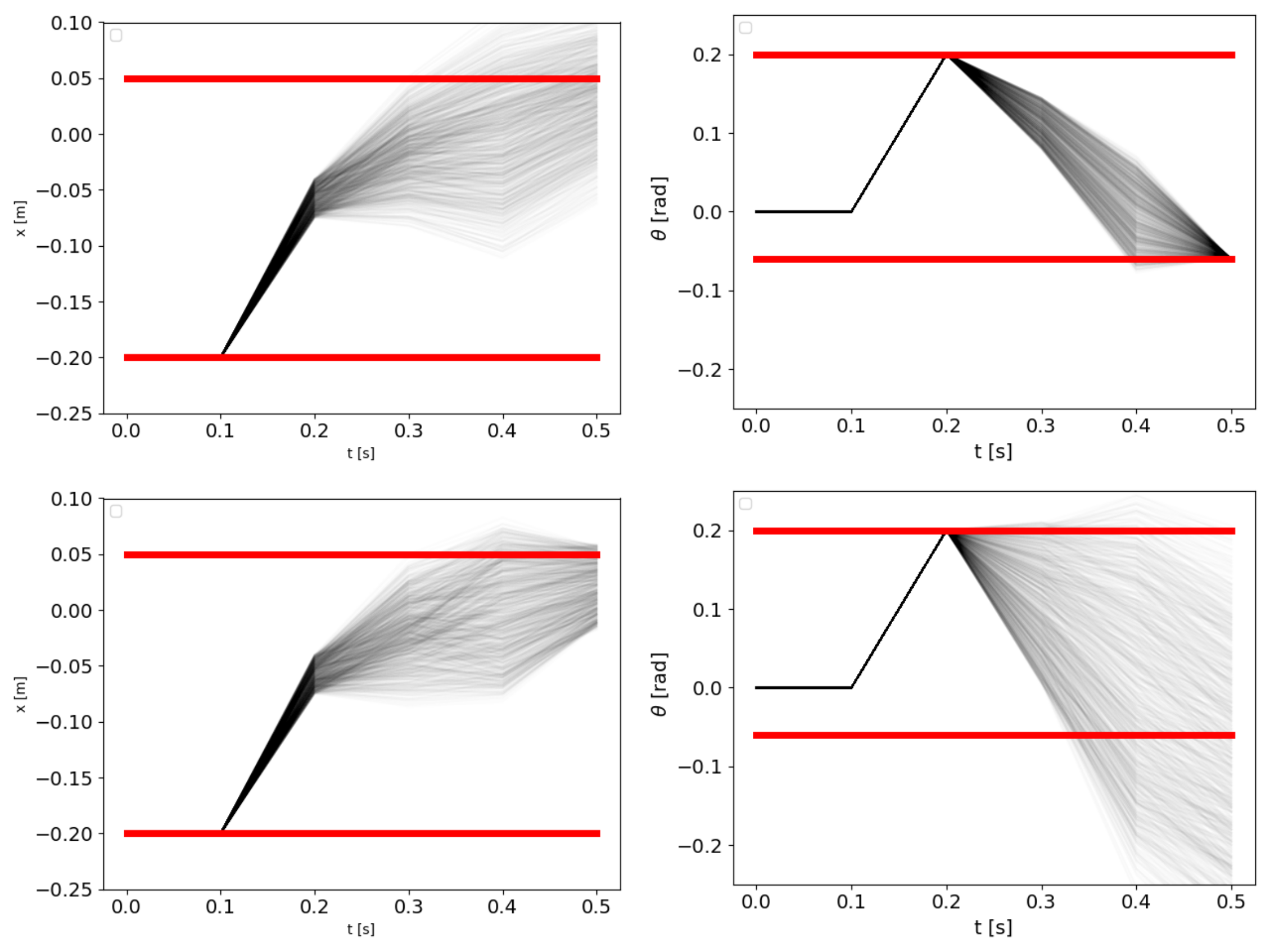} %
    \caption{Simulated trajectories for cartpole system using our closed-loop controller. Top: $\Delta=0.6$ and $\Delta_\text{test} = 0.510$, bottom: $\Delta=0.2$ and $\Delta_\text{test} = 0.188$, where $\Delta$ is input of optimization and $\Delta_\text{test}$ is the empirically obtained success rate from MC simulation. Red lines show boundaries specified in chance constraints.}
    \label{fig:diff_chance_cartpole}
    \vspace{-2em}
\end{figure}

The results  using ERM and our controller for the open-loop trajectory are shown in \fig{fig:cartpole_ERM}, \fig{fig:cartpole_open}. We observed that the proposed open-loop controller shows the better satisfaction of chance constraints compared to the ERM-based method. This is because our method explicitly considers propagation of uncertainty for SDLCS while the ERM-based method is unable to consider. Also, we observe that the gap between the commanded $\Delta$ used in our optimization and $\Delta_\text{test}$ obtained from MC simulation over testing dataset is smaller the gap between the commanded $\Delta$ used in ERM method and $\Delta_\text{test}$ obtained from MC simulation over testing dataset. Again this is because our method could capture the evolution of uncertainty for SDLCS.
However, even our open-loop controller does not show the much better performance than the ERM. To show the higher $\Delta_\text{test}$, we need to input the higher $\Delta$ as an input of optimization. It is quite difficult especially for long-horizon planning problems since uncertainty keeps evolving, which can be observed from both \fig{fig:cartpole_ERM} and \fig{fig:cartpole_open}. 

Next, we discuss the difference among our proposed contact-aware closed-loop, the non-contact-aware (i.e., $L_k=0, \forall k$ in \eq{feedback}) closed loop, and the open-loop controllers. 
% To observe how controllers behave in a longer horizon, we set $T = 20$ and use the same values for the other parameters.
We observed that  in \tab{open_closed_vio_comarison}, \eq{equation_control_bilevel_MPCC} for  open-loop controller with high $\Delta$ was unable to find feasible solutions but \eq{equation_control_bilevel_MPCC} for  closed-loop controller could find feasible solutions. Since the closed-loop controller can change feedback gains to satisfy chance constraints, it could find feasible solutions with high $\Delta$. Also, \tab{open_closed_vio_comarison} shows that the contact-aware closed-loop controller could find the feasible solution with high $\Delta = 0.8, 0.7$  but the non-contact-aware controller (i.e., $L_k=0, \forall k$ in \eq{feedback}) could not. For SDLCS, introducing feedback to both states and forces is important to realize the robust motion. 
% Therefore, we conclude that introducing feedback to both states and forces is important to design feedback controller over SDLCS. 
The MC simulation results using our contact-aware closed-loop controller are shown in \fig{fig:diff_chance_cartpole}. In contrast to \fig{fig:cartpole_ERM} and  \fig{fig:cartpole_open}, the closed-loop controller could bound the distribution of the states because it controls covariance. 
% Also, we observed that our closed-loop controller could realize much higher $\Delta$, which can be also verified in \tab{open_closed_vio_comarison}. 
% In \tab{open_closed_vio_comarison}, \eq{equation_control_bilevel_MPCC} for  open-loop controller with high $\Delta$ was unable to find feasible solutions but \eq{equation_control_bilevel_MPCC} for  closed-loop controller could find feasible solutions. Since the closed-loop controller can change feedback gains to satisfy chance constraints, it could find feasible solutions with high $\Delta$.

We discuss computational results. 
Firstly,  we observe that our important-particle method converges and the gap between $\Delta_\text{train}$ and $\Delta_\text{test}$ is small once it finishes its third time iteration. It means that our important-particle method could successfully find feasible trajectories with relative small number of particles. 
Secondly,   in \tab{table_cartpole} the  important-particle method shows the higher $\Delta_\text{train}$ as the number of particles used in optimization increases. The proposed important-particle method shows better convergence (in total 208 s to have $\Delta_\text{train}\geq 0.49$) than the naive  method (620 s with 50 particles to have $\Delta_\text{test}\geq 0.49$) since our important-particle method keeps choosing the worst-case particles which break chance constraints. 

% Based on these observations, we verified that our controller could successfully generate robust trajectories under SDLCS. 
% 
% \tab{open_closed_vio_comarison} summarizes 

\begin{table}[t]
    \caption{{Comparison of feasibility for cartpole system among open-, non-contact-aware closed, and contact-aware-closed controllers with different $\Delta$. $\circ$ and $\times$ show if optimization finds a feasible solution or not, respectively.}}
    \centering
    \begin{tabular}{c|c|c|c|c|c}
    $\Delta$ & $0.8$& $0.7$ & $0.6$ & $0.4$  & $0.2$ \\
         \hline
         Open-loop  & $\times$ & $\times$ & $\times$ & $\circ$ & $\circ$ \\
         \hline
         Non-contact-aware closed-loop & $\times$ & $\times$ & $\circ$ & $\circ$ & $\circ$ \\
                  \hline
         Contact-aware closed-loop & $\circ$ & $\circ$ & $\circ$ & $\circ$ & $\circ$
    \end{tabular}
    \label{open_closed_vio_comarison}
\end{table}

\begin{table}[t]
    \caption{{Comparison of safe probability and runtime for cartpole system between important-particle  method (top) with $\gamma = 10, \eta=10$ and naive method (bottom) with $\Delta = 0.6$ for designing the closed-loop controller. $T$ represents runtime for each iteration and $n_p$ is the number of particles.}}
    \centering
    \begin{tabular}{c|c|c|c}
     \text{iter} & $1$& $2$ & $3$ \\
         \hline
         $\Delta_{\text{train}}$  & 0.2708 & 0.09 & 0.592 \\
         \hline
         $\Delta_{\text{test}}$  & N/A & N/A & 0.588 \\
         \hline
         T [s]  & 25 & 35 & 148
         \\
         \hline
         $n_p$  & 10 & 20 & 30
    \end{tabular}
        \begin{tabular}{c|c|c|c|c}
     Case & 1& 2 & 3 & 4\\
         \hline
         $\Delta_{\text{test}}$ & 0.277 & 0.376 & 0.451 & 0.499 \\ 
         \hline
         T [s] & 25 & 26 & 55 & 620
         \\ 
         \hline
         $n_p$ & 10 & 20 & 30 & 50
    \end{tabular}
    \label{table_cartpole}
\end{table}
% want to say:
% open loop is not suited for long-term planning
% ERM is worse than TO

% 

\subsection{Acrobot with Soft Joints}
\begin{figure}[t]
    \centering
    \includegraphics[width=0.48\textwidth]{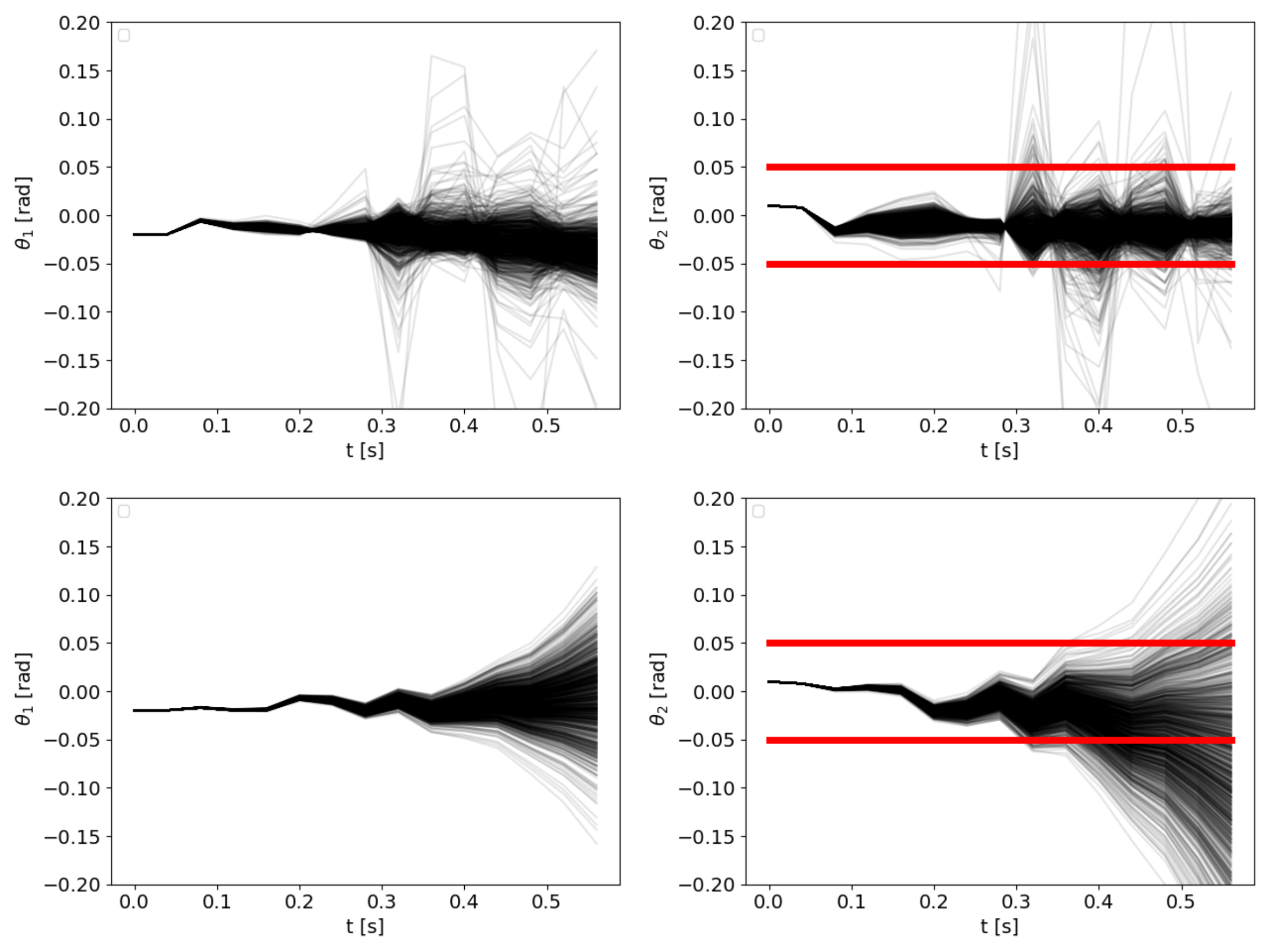} %
    \caption{Simulated trajectories for acrobot using our open- and closed-loop controllers.  Top: closed-loop controller with $\Delta=0.8$ and $\Delta_\text{test} = 0.771$, bottom: open-loop controller with $\Delta=0.4$ and $\Delta_\text{test} = 0.366$. Red lines show boundaries specified in chance constraints. The reader should note that open-loop controller solution was infeasible for $\Delta=0.8$, and thus we show results for $\Delta=0.4$.}
    \label{fig:acrobot_closed}
\end{figure}

\begin{table}[t]
    \caption{{Comparison of safe probability and runtime for acrobot system between important-particle method (top) with $\gamma = 4, \eta=4$ and naive method (bottom) with $\Delta = 0.8$.} $T$ represents runtime for each iteration and $n_p$ is the number of particles.}
    \centering
    \begin{tabular}{c|c|c|c|c|c|c|c}
     \text{iter} & $1$& $2$ & $3$ & $4$& $5$ & $6$ & $7$ \\
         \hline
         $\Delta_{\text{train}}$  & 0.426 & 0.485 & 0.562 & 0.625 & 0.363 & 0.593  & 0.763 \\
                %   $\Delta_{\text{train}}$  & 0.4268 & 0.4851 & 0.5628 & 0.6256 & 0.3633 & 0.5937  & 0.7639 \\
         \hline
         $\Delta_{\text{test}}$  & N/A & N/A & N/A & N/A & N/A & N/A & 0.771 \\
         \hline
         $T$ [s]  & 31 & 97 & 557 & 887 & 698 & 2450 & 779
         \\
         \hline
        $n_p$  & 4 & 8 & 12 & 16 & 20 & 24 & 28 
    \end{tabular}
        \begin{tabular}{c|c|c|c|c|c|c|c}
     Case & 1& 2 & 3 & 4 & 5 & 6 & 7\\
         \hline
         $\Delta_{\text{test}}$ & 0.009 & 0.103 & 0.159 & 0.541 & 0.670 & 0.553 & 0.539 \\ 
         \hline
         $T$ [s] & 31 & 15 & 229 & 260 & 944 & 3993 & 901 
         \\ 
         \hline
        $n_p$  & 4 & 8 & 12 & 16 & 20 & 24 & 28
    \end{tabular}
    \label{table_acrobot}
\end{table}

We also demonstrate our controller for acrobot with soft joints system (see \cite{9197568} for more details). $\theta_{1}$ is the first joint angle and $\theta_{2}$ is the second joint angle. 
$u_{1}$ is the control at the second joint and $\lambda_{1}, \lambda_{2}$ are the reaction forces at from the wall 1, 2, respectively.
We have the following deterministic physical parameters.$g=9.81$ is the gravitational acceleration, $m_1=0.5, m_2=1.0$ are the mass of the  pole, cart, respectively. $l_1=0.5$ is the length of the rod from the first to the second joint. $d=0.2$ is the angle limit of $\theta_1$.  
We consider the stochastic physical parameters $k$ and $l_2$ where $k$ is the stiffness of the walls and $l_2$ is the length of the second rod. We assume that $k$ follows uniform distribution where the upper bound and the lower bound of the distribution is 1.6 and 0.6, respectively. We assume that $l_2$ follows a truncated Gaussian distribution where we set the mean to 1.0, variance to 0.01, the upper bound of the interval is 1.3, and the lower bound of the interval is 0.7, respectively.
We set $dt=0.04$ for the explicit Euler integration and $T=15$.

The open- and closed-loop trajectories are shown in \fig{fig:acrobot_closed}. We observed that both controller could satisfy chance constraints over the testing dataset and the closed-loop controller shows the better performance. \tab{table_acrobot} shows that the important-particle method shows the higher $\Delta_\text{test} = 0.771$ than the naive method with the same number of particles used in optimization.

% \begin{table}[t]
%     \caption{{Comparison of feasibility for acrobot system between contact-aware and non-contact-aware closed-loop controllers with different $\Delta$.}}
%     \centering
%     \begin{tabular}{c|c|c|c|c}
%     $\Delta$ & $0.8$& $0.7$ & $0.6$  & $0.5$ \\
%          \hline
%          Contact-aware  & feasible & feasible & feasible & feasible \\
%          \hline
%          Non-contact-aware & infeasible & infeasible & feasible & feasible 
%     \end{tabular}
%     \label{contact-aware_vio_comarison}
% \end{table}

% \begin{figure}[t]
%     \centering
%     \includegraphics[width=0.499\textwidth]{figures/acrobot_contact_gain_comparison.png} %
%     \caption{Simulated trajectories for acrobot using our contact-aware and non-contact-aware controllers.  Top: contact-aware controller with $\Delta=0.7$ and $\Delta_\text{test} = 0.630$, bottom: non-contact-aware controller with $\Delta=0.6$ and $\Delta_\text{test} = 0.587$.}
%     \label{fig:acrobot_contact_gain_discussion}
% \end{figure}

% \subsection{Computation Results}
\section{Discussion}\label{sec:discussion}
Stochastic complementarity systems are not well understood in literature.
This paper presents a study of SDLCS to perform covariance steering using particles. Under the assumption of uniqueness of trajectory ($\bar{F}$ is P-matrix) for complementarity systems, the proposed method is able to compute covariance controller for SDLCS. We presented an important-particle method to alleviate computational complexity of the resulting optimization problem. It is shown that our work could design open- and closed-loop controllers with chance constraints by appropriately considering the evolution of uncertainty for SDLCS.

In the future, we would like to study more general manipulation systems by relaxing the assumption on uniqueness of trajectory for SDLCS. Another limitation of this work is that the computation is still demanding. Thus, we would like to employ distributed optimization techniques such as ADMM \cite{boyd2011distributed}.
% One of the limitations of this work is that i
% Future work would be 
% There are many exciting future directions. Firstly, this work assumes that $F$ is P-matrix.
% Secondly, there would be a richer controller structure
% Thirdly, learning-rate with Neural net?

%%%%%%%%%%%%%%%%%%%%%%%%%%%%%%%%%%%%%%%%%%%%%%%%%%%%%%%%%%%%%%%%%%%%%%%%%%%%%%%%

\bibliographystyle{IEEEtran}
\bibliography{main}

\end{document}